\documentclass[runningheads]{llncs}
\usepackage{graphicx}
\usepackage{algorithm}
\usepackage{algorithmic}
\usepackage{multirow}
\usepackage{makecell}
\usepackage{rotating}
\usepackage{amssymb}
\usepackage{amsmath}

\begin{document}
\title{Enhancing Visual Dialog State Tracking through Iterative Object-Entity Alignment in Multi-Round Conversations}
\titlerunning{MDST by Object-Entity Alignment in Visual Dialog}
%
\author{Wei Pang\inst{1} \and
Ruixue Duan\inst{1} \and
Jinfu Yang\inst{2} \and 
Ning Li\inst{1}}
\authorrunning{Wei Pang et al.}
%
\institute{Beijing Information Science and Technology University, Beijing, China
\email{\{pangweitf,duanruixue\}@bistu.edu.cn}\\
\and
Beijing University of Technology, Beijing, China\\}
\maketitle              
\begin{abstract}
Visual Dialog (VD) is a task where an agent answers a series of image-related questions based on a multi-round dialog history. However, previous VD methods often treat the entire dialog history as a simple text input, disregarding the inherent conversational information flows at the round level. In this paper, we introduce Multi-round Dialogue State Tracking model (MDST), a framework that addresses this limitation by leveraging the dialogue state learned from dialog history to answer questions. MDST captures each round of dialog history, constructing internal dialogue state representations defined as 2-tuples of vision-language representations. These representations effectively ground the current question, enabling the generation of accurate answers. Experimental results on the VisDial v1.0 dataset demonstrate that MDST achieves a new state-of-the-art performance in generative setting. Furthermore, through a series of human studies, we validate the effectiveness of MDST in generating long, consistent, and human-like answers while consistently answering a series of questions correctly.

\end{abstract}

\section{Introduction}
Vision-language based multi-modal tasks have gained significant attention at the intersection of computer vision and natural language processing. Tasks such as Visual Question Answering (VQA) \cite{VQA}, and Visual or Video Dialogue \cite{GuessWhat,VD17,VGDS} require the fusion of visual and textual information. Among these tasks, Visual Dialog (VD) \cite{VD17} poses a unique challenge that goes beyond simple question answering grounded in an image. VD involves comprehending conversational language, navigating through multi-round dialog history, and reasoning based on visual and textual contents to generate coherent answers.

While previous methods in VD have made progress, they often overlook the inherent information flows and round-level interactions within the dialog history. Existing models \cite{desai2018visdialch,DualVD,LTMI,VDBERT,Pretrained,LTMI-LG,ICMU,UTC} commonly concatenate the entire dialog history into a single text sequence, lacking explicit focus on the most relevant history clues. Although attention mechanisms, such as sequential attention \cite{HCIAE}, co-attention \cite{MCA,HACAN}, dual-attention\cite{DMRM}, and multi-view attention \cite{MVAN}, have been proposed, they still treat each round of the dialog history independently.

To address these limitations, we propose the multi-round dialogue state tracking model (MDST) for Visual Dialog. Unlike prior work, MDST explicitly models the round-level interactions in dialog history. We define the dialogue state in VD as a 2-tuple of vision and language states, where vision states capture object-level representations and language states represent dialog entity-level representations.

In MDST, each round question from the dialog history is processed, grounding the question in the dialogue state to yield question-guided visual-textual clues. These clues are then used to decode accurate answers, while updating the dialogue states accordingly. Notably, vision states remain unchanged throughout the dialogue, while language states are updated in each round. We align the vision and language representations in dialogue states in an object-entity fashion, facilitating the grounding of follow-up questions.

Experimental results on the VisDial v1.0 dataset demonstrate that our proposed model achieves state-of-the-art performance in generative setting. Further examinations reveal that MDST consistently answers questions correctly, with a joint answer accuracy (JACC) of 79.8\% in the generative setting. Moreover, MDST generates human-like responses, as validated through human studies. To summarize, our contributions are three-fold:
\begin{itemize}
\item We propose a novel multi-round dialogue state tracking model (MDST) for Visual Dialog. The MDST, including representations of image objects and representations of dialog entities, models the inherent interactions in dialog history at the round level.

\item We achieve new state-of-the-art results on most evaluation metrics on VisDial v1.0, and find that the alignment of vision-language in dialogue states could improve the final performance significantly.

\item We introduce JACC to evaluate the answer quality, and find that our MDST can continuously generate correct answers as proved by JACC of 79.8\% that means about 8 rounds in 10 are correct on VisDial v1.0 val.
\end{itemize}

\section{Related Work}
Dealing with dialog history as a simple text input in Visual Dialog (VD) has been a prevailing practice since the works of LF \cite{VD17}, HCIAE \cite{HCIAE}, and LTMI \cite{LTMI}. However, recent research has highlighted the limitations of such approaches in explicitly capturing round-level interactions between dialog rounds \cite{GST,VDST}. Existing work can be categorized into three groups based on their handling of dialog history.

Firstly, attention-based models \cite{VD17,HCIAE,CoAtt,RvA,DMRM,ReDAN,DAN,SeqDialN,MVAN,MCA} typically encode each round of history separately to obtain a set of history embeddings. Sequential attention is applied in HCIAE \cite{HCIAE} to attend to the history and image sequentially. MCA \cite{MCA} leverages modular co-attention to fuse visual and textual modalities. DMRM \cite{DMRM} utilizes dual-attention mechanisms to resolve textual and visual co-references. MVAN \cite{MVAN} introduces multi-view attention to fuse the question and history at both the sentence and word level.

Secondly, graph-based models \cite{KBGN,GoG,HKNet} construct a graph representation of the entire dialog history, where each node represents a question-answer (QA) pair. KBGN \cite{KBGN} and LTMI-GoG \cite{GoG} establish edges between nodes to indicate coreference relations between QA pairs. However, these graph-based approaches can suffer from scalability issues as the graph size grows with the dialogue.

Thirdly, concatenation-based models treat the entire dialog history as a single sentence. DualVD \cite{DualVD} packs the dialog history into a long string encoded by an LSTM. UTC \cite{UTC}, ICMU \cite{ICMU}, and LTMI \cite{LTMI} concatenate each QA pair as a text sequence, separated by a special token (e.g., [SEP]), and input them into a transformer encoder.

In summary, prior approaches for handling dialog history in VD have not explicitly modeled interactions at the round level of granularity. This limitation hinders their ability to capture the nuanced dynamics of multi-round dialogues.

\section{Model}

\begin{figure*}[!htb]
\centering
\includegraphics[width=0.95\textwidth,scale=0.8, clip=true,keepaspectratio]{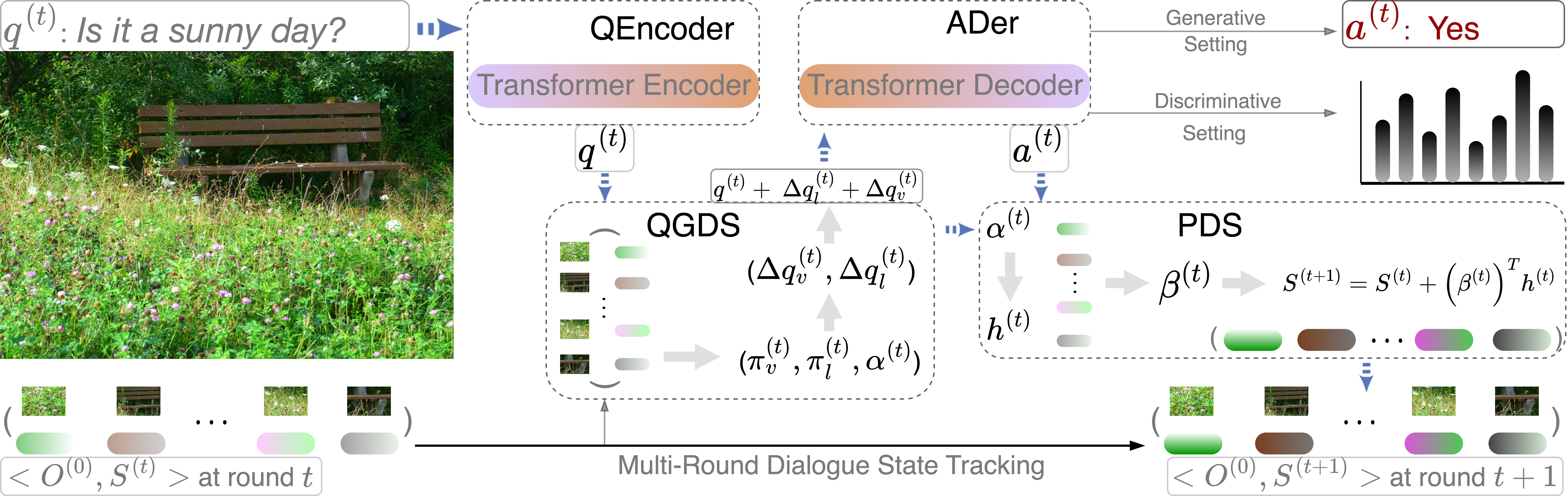}
\caption{Overall structure of our proposed MDST model for Visual Dialog.}
\label{fig_overview}
\end{figure*}

Figure~\ref{fig_overview} presents an overview of our Multi-Round Dialogue State Tracking (MDST) model, which consists of four main modules: Question Encoder (QEncoder), Question Grounding on Dialogue State (QGDS), Answer Decoder (ADer), and Postdiction on Dialogue State (PDS). In the remainder of this section, we go into more detail on each module.

\noindent {\bf Problem Formulation}
Given an image $I$ and a multi-round dialogue history $H = {C, (q^{(1)}, a^{(1)}), \ldots, (q^{(t-1)}, a^{(t-1)})}$ up to round $t-1$, where $C$ represents the image caption and $(q^{(\cdot)}, a^{(\cdot)})$ denotes the previously experienced question-answer pairs, the dialogue agent aims to respond to the current question $q^{(t)}$ at round $t$. This response involve generating a free-form natural language answer in a generative setting.

We first extract object-level image features using Faster-RCNN \cite{RCNN}. For each image, we select the top-ranked $N$ objects, each of them is of size 2048-dim and projected to low-dimension features with size $d$ through a linear layer as:
{\setlength{\abovedisplayskip}{0.02cm}
\setlength{\belowdisplayskip}{0.0cm}
\begin{small}\begin{align}
\label{equ_Img_F} O^{f} &= \mathrm{RCNN}(I),\\ 
\label{equ_Img_O} O^{(0)} &= \mathrm{LayerNorm}(\mathrm{ReLU}(W_{o}O^{f} + b_{o})),
\end{align}\end{small}}
\noindent where LayerNorm is the layer normalization \cite{LayerNorm}, $W_{o}$ and $b_{o}$ are learnable parameters, $O^{(0)}\in \mathbb{R}^{N\times d}$ represents a set of object-level image features. Furthermore, we insert two special pseudo-object features: NULL ($\epsilon$) and ALL ($\chi$), where NULL is a zero vector of size d, and ALL denotes the representation of the whole image by taking the mean of $O^{(0)}$. Thus, we get a new set of $N+2$ object features of $O^{(0)}\in \mathbb{R}^{(N+2)\times d} = O^{(0)}{\ }\cup{\ }\{\epsilon\}{\ }\cup{\ }\{\chi\}$. For clarity, we omit all the biases in the remainder of this section.

Let ${<}O^{(0)}, S^{(0)}{>}$ denote the initial dialogue state, where $S^{(0)} \in \mathbb{R}^{(N+2) \times d}$ is initialized as a set of zero vectors of the same size. In our approach, the image caption is treated as the zeroth round QA pair $C^{(0)}$, which serves as the initialization for $S^{(0)}$ at the beginning of the dialogue.

\noindent {\bf Question Encoder (QEncoder)}
For encoding both the question and the image caption, we employ a standard Transformer encoder \cite{Transformer}. This encoder generates contextual representations, denoted as $q^{(t)}$ and $C^{(0)}$, where $l$ represents the length of the question or caption. It is worth noting that, for simplicity, we use the same symbol to represent both the textual string and its corresponding representation.

\noindent {\bf Question Grounding on Dialogue State (QGDS)}
QGDS aims to ground current question $q^{(t)}$ in dialogue state, yielding question-related textual clues on language states and visual clues on vision states. To better associate question with vision and language states, three probability distributions are designed: word-entity alignment between question words and language states, word-object alignment between question words and vision states, and switching probability. Before going into detail, we introduce a notation to express a non-linear transformation layer, to which dropout regularization and layer normalization are applied:

{\setlength{\abovedisplayskip}{0.02cm}
\setlength{\belowdisplayskip}{0.0cm}
\begin{small}\begin{align} \label{equ_mlp}
\mathrm{MLP}(x) = \mathrm{LayerNorm}(\mathrm{Dropout}(\mathrm{GELU}(Wx))),
\end{align}\end{small}}
\noindent where $x$ is the input: a vector or a matrix, with learnable weight $W$ of the size varying with the input.

Because many textual relations (e.g., co-reference) existing in question and previous history \cite{VD17}, our model will associate current question words with its most related dialog entities in language states using a learnable word-entity alignment distribution $\pi^{(t)}_{l}\in \mathbb{R}^{l\times N}$ in Eq.\ref{equ_pi_l}. To ground current question in an image, we then calculate a cross-modal matching as in Eq.\ref{equ_pi_v},

{\setlength{\abovedisplayskip}{0.02cm}
\setlength{\belowdisplayskip}{0.0cm}
\begin{small} \begin{align}
\label{equ_pi_l} \pi^{(t)}_{l} &= \mathrm{softmax}(\mathrm{MLP}(q^{(t)})\cdot \mathrm{MLP}(S^{(t)})^{T}/\sqrt{d}),\\
\label{equ_pi_v} \pi^{(t)}_{v} &= \mathrm{softmax}(\mathrm{MLP}(q^{(t)})\cdot \mathrm{MLP}(O^{(0)})^{T}/{\sqrt{d}}),
\end{align}\end{small}}
\noindent where $\pi^{(t)}_{v}\in \mathbb{R}^{l\times N}$ represents word-object alignment distribution between question words and objects in vision states.

Switching probability is designed to 1) determine whether current question is related to previous dialog history; 2) provide a weight to fuse two alignment distributions because there is one-to-one correspondence (i.e., object-entity) between vision and language states:

{\setlength{\abovedisplayskip}{0.02cm}
\setlength{\belowdisplayskip}{0.0cm}
\begin{scriptsize}\begin{align} \label{equ_alpha_phi}
\varphi^{(t)} &= \mathrm{sigmoid}(\frac{w(\mathrm{MLP}(q^{(t)})\mathrm{MLP}(S^{(t)})^{T})\mathop{.mean}\limits_{l\to 1}}{\sqrt{N+2}}),
\end{align}\end{scriptsize}}

\noindent where $\mathop{.mean}$ takes the mean on the $l$ dimension with trainable parameter $w\in \mathbb{R}^{(N+2)\times 1}$. $\varphi^{(t)}\in [0,1]$ is a weight measured the relationship between question and dialog history. The larger value of $\varphi^{(t)}$, the less relevant current question is to dialog history. Experiment shows introducing $\varphi^{(t)}$ can contribute to better the final performance.

The question-guided textual context $\Delta q^{(t)}_{l}$ is obtained by a weighted sum of language stats over both word-entity and word-object alignment distributions, as denoted in Eq.\ref{equ_delta_l}:

{\setlength{\abovedisplayskip}{0.02cm}
\setlength{\belowdisplayskip}{0.0cm}
\begin{small}\begin{align}
\label{equ_delta_l} \Delta q^{(t)}_{l} &= S^{(t)}(\pi^{(t)}_{l} + \varphi^{(t)}\pi^{(t)}_{v}),\\
\label{equ_delta_v} \Delta q^{(t)}_{v} &= O^{(0)}(\pi^{(t)}_{v} + (1 - \varphi^{(t)})\pi^{(t)}_{l}),
\end{align}\end{small}}
\noindent where $\Delta q^{(t)}_{l} \in \mathbb{R}^{l\times d}$ represents history context relevant to current question composed of two parts. The first part consists of an explicit history attention directly from question to language states, while the second part contains an aligned history attention indirectly from question, via vision states, to language states, weighted by switching probability. Similarly, the question-guided visual context is written as in Eq.\ref{equ_delta_v}, where $\Delta q^{(t)}_{v} \in \mathbb{R}^{l\times d}$ represents the focused visual regions relevant to current question from two parts, including an explicit visual attention from question to  vision states and an implicit ones via the cross-modal alignment.

Finally, we use the sum of the three components to denote the final question representation as in $q^{(t)} + \Delta q^{(t)}_{l} + \Delta q^{(t)}_{v}$, which is decoded in next ADer module.

\noindent {\bf Answer Decoder (ADer)}
In ADer, we utilize a standard Transformer decoder as the backbone for the generative setting. It takes the final question representation, obtained by combining the question representation $q^{(t)}$ with the question-guided textual context $\Delta q^{(t)}_{l}$ and the question-guided visual context $\Delta q^{(t)}_{v}$, as input. The decoder autoregressively generates the next word one by one until it encounters an end-of-sequence token, producing a free-form natural answer. Formally, the ADer module can be expressed as:
{\setlength{\abovedisplayskip}{0.02cm}
\setlength{\belowdisplayskip}{0.0001cm}
\begin{small}\begin{align} \label{equ_ADer}
a^{(t)} &= \mathrm{Decoder}(q^{(t)} + \Delta q^{(t)}_{l} + \Delta q^{(t)}_{v}),
\end{align}\end{small}}
\noindent where $a^{(t)}$ represents the output of the decoder, which not only represents the free-form natural answer of length $\ell$, but also denotes its contextualized representations over the words: $a^{(t)}\in \mathbb{R}^{\ell \times d}$. The decoder progressively generates each word based on the input representation.

In the discriminative setting, we encode each of the 100 candidate answers using another Transformer encoder. We score these candidate answers by computing the dot product similarity with the final question representation. The candidate answer with the highest score is selected as the response.

\noindent {\bf Postdiction on Dialogue State (PDS)}
PDS is responsible for updating the representation of previously experienced language states with the new question-answer (QA) pair. This module incorporates the new QA pair as new information into the dialogue history, refining the dialogue state representation. It's worth noting that the language states are updated from $S^{(0)}$ to $S^{(1)}$ with the image caption at the beginning, while the vision states remain unchanged throughout the dialogue.

The PDS module leverages an alignment distribution to fuse the QA pair and capture word-word interactions between the question and answer. The alignment distribution is defined as follows:
{\setlength{\abovedisplayskip}{0.02cm}
\setlength{\belowdisplayskip}{0.0001cm}
\begin{small}\begin{align}\label{equ_QAPair}
\nonumber
\alpha^{(t)}&=\mathrm{softmax}(\mathrm{MLP}(q^{(t)}+\Delta q^{(t)}_{l})\mathrm{MLP}(a^{(t)})^{T}/{\sqrt{d}}),\\
h^{(t)} &= q^{(t)} + \alpha^{(t)} a^{(t)},
\end{align}\end{small}}
\noindent where $\alpha^{(t)}\in \mathbb{R}^{l\times \ell}$ represents the word-word alignment distribution between the question and answer. $h^{(t)}\in \mathbb{R}^{l\times d}$ denotes the final representation of the QA pair, which is used to update the previous language states, as follows:
{\setlength{\abovedisplayskip}{0.02cm}
\setlength{\belowdisplayskip}{0.0001cm}
\begin{small}\begin{align}\label{equ_Update}
\nonumber
\beta^{(t)} &= \mathrm{softmax}(\mathrm{MLP}(h^{(t)}+\Delta q^{(t)}_{l})\mathrm{MLP}(S^{(t)})^{T}/{\sqrt{d}}),\\
S^{(t+1)} &= S^{(t)} + (\beta^{(t)})^{T} h^{(t)},
\end{align}\end{small}}
\noindent where $\beta^{(t)}\in \mathbb{R}^{l\times (N+2)}$ represents the assignment probability, indicating how the new QA information is distributed among the language states. It provides a word-entity alignment distribution for associating the new information with the language states. The language states are then updated to $S^{(t+1)}$ by adding the assigned new information to $S^{(t)}$. The updated dialogue states ${<}O^{(0)},S^{(t+1)}{>}$ are used as input for the Question Grounding on Dialogue State (QGDS) module in the next round, and this process continues iteratively.

\section{Experiment}

\noindent {\bf Datasets and Evaluation}
We conduct our experiments on the VisDial v1.0 dataset, which consists of a standard train/val/test split. To evaluate the performance, we employ NDCG and retrieval metrics, including MRR, Mean rank, and R@{1, 5, 10}, following the conventions of previous studies \cite{ICMU,UTC}. Additionally, we assess the quality of generated answers by generating 2064 dialogues for 10 rounds on the VisDial v1.0 validation set and calculating the following metrics: Joint Answer Accuracy (JACC) measures the percentage of correct QA pairs among all the generated QA pairs. It assesses whether the generated answers are correct given the corresponding images. Average Answer Length (AvgLen) calculates the average length of generated answers.

\noindent {\bf Implementation Details}
We utilize a Transformer encoder-decoder architecture as the backbone. The encoder and decoder consist of 12 layers with 12 heads and 768 hidden states, respectively. For image features, we extract bottom-up features of 36 objects using Faster-RCNN. We employ the Adamax optimizer with an initial learning rate of 1e-3, which linearly decreases to 5e-5 following a scheduled warmup of 0.2. The model is trained for 20 epochs with a batch size of 32. The word embeddings, shared between encoders and decoders, are set to 768 dimensions. For training, we compute the negative log-likelihood of ground-truth and generated answer.

\noindent {\bf Comparison to State-of-the-Art Methods}
We compare our approach to several state-of-the-art methods, categorizing them based on how they utilize the dialog history: 1) Attention-based models: MN \cite{VD17}, HCIAE \cite{HCIAE}, CoAtt \cite{CoAtt}, DAM \cite{DAM}, DMRM \cite{DMRM}, ReDAN \cite{ReDAN}, SeqIPN \cite{SeqDialN}, MVAN \cite{MVAN}, and MCA \cite{MCA}. 2) Graph-based models: KBGN \cite{KBGN}, LTMI-GoG \cite{GoG}, and HKNet \cite{HKNet}. 3) Concatenation-based models: LateFusion \cite{desai2018visdialch}, DualVD \cite{DualVD}, LTMI \cite{LTMI}, LTMI-LG \cite{LTMI-LG}, VDBERT \cite{VDBERT}, Visdial-BERT \cite{Pretrained}, ICMU \cite{ICMU}, and UTC \cite{UTC}. 4) Dialogue State Tracking (DST) based model: Our MDST model. It's important to note that our MDST model for VisDial v1.0 is trained from scratch, without relying on pretraining or fine-tuning on additional large-scale datasets.

\begin{table}[!htb]\small
\centering
\caption{Comparisons on VisDial v1.0 val in the generative setting.} \label{tab_gen}
\setlength{\tabcolsep}{2.8mm}{
\begin{tabular}{l|ccccc|c}
\hline
Model& MRR$\uparrow$ & R@1$\uparrow$ & R@5$\uparrow$ & R@10$\uparrow$ & Mean$\downarrow$ & NDCG$\uparrow$\\
\hline
LateFusion&46.57&36.20&56.40&63.40&19.44&54.21\\
MN &47.83&38.01&57.49&64.08&18.76&56.99\\
HCIAE &49.07&39.72&58.23&64.73&18.43&59.70\\
CoAtt &49.64&40.09&59.37&65.92&17.86&59.24\\
DAM  &50.51&40.53&60.84&67.94&16.65&60.93\\
DMRM &50.16&40.15&60.02&67.21&15.19&-\\
ReDAN &49.60&39.95&59.32&65.97&17.79&59.41\\
SeqIPN&47.86&38.16&57.08&64.89&15.27&60.72\\
SeqMRN&49.22&38.75&59.62&68.47&\textbf{13.00}&63.01\\
\hline
SKANet &45.53&36.17&55.05&61.41&19.79&-\\
KBGN &50.05&40.40&60.11&66.82&17.54&60.42\\
LTMI-GoG &51.32&41.25&61.83&69.44&15.32&62.63\\
\hline
LTMI &50.38&40.30&60.72&68.44&15.73&61.61\\
LTMI-LG &51.30&41.34&61.61&69.06&15.26&63.23\\
LTMI-LG$^{\ast}$ &51.43&41.68&61.96&\textbf{69.87}&14.89&63.53\\
UTC &52.22&\textbf{42.56}&62.40&69.51&15.67&63.86\\
\hline
MDST (Ours)&\textbf{53.49}&\textbf{42.56}&\textbf{62.47}&69.77&14.94&\textbf{65.03} \\
\hline
\end{tabular}}
\end{table}
\begin{table}[!htb]\small
\centering
\caption{Main comparisons on VisDial v1.0 test.} \label{tab_test_gen}
\setlength{\tabcolsep}{2.8mm}{
\begin{tabular}{l|ccccc|c}
\hline
Model& MRR$\uparrow$ & R@1$\uparrow$ & R@5$\uparrow$ & R@10$\uparrow$ & Mean$\downarrow$ & NDCG$\uparrow$\\
\hline
LTMI-LG &64.0&50.63&80.58&90.20&4.12&58.55\\
LTMI &64.08&50.20&80.68&90.35&4.05&59.03\\
VDBERT&65.44&51.63&82.23&90.68&3.90&59.96\\
LTMI-GoG &63.52&50.01 &80.13&89.28&4.31&61.04 \\
ICMU &\textbf{66.82}&53.50&83.05&92.05&3.59&61.30\\
UTC &66.27&52.25 &83.55&92.23&\textbf{3.48}&62.65\\
\hline
MDST (Ours)&66.78&\textbf{53.58}&\textbf{83.69}&\textbf{92.62}&3.54&\textbf{63.18} \\
\hline
\end{tabular}}
\end{table}

\begin{table}[!htb]\small
\centering
\caption{Ablation study on VisDial v1.0 val in the generative setting.} \label{tab_ablation}
\setlength{\tabcolsep}{2.8mm}{
\begin{tabular}{ll|ccccc|c}
\hline
\#&Model{\ }& MRR & R@1 & R@5 & R@10  & Mean & NDCG \\
\hline
1&MDST&{53.49}&{42.56}&{62.47}&69.77&14.94&{65.03} \\ \hline
2&$\quad$-QGDS-PDS &50.80&40.55&60.79&67.46&15.73&61.15\\
3&$\qquad$-$\alpha^{(t)}$&52.79&41.65&61.43&69.05&15.20&63.64\\
4&$\quad$-NULL-ALL &53.27&41.91&61.97&69.34&15.06&64.48\\
\hline
\end{tabular}}
\end{table}

Table~\ref{tab_gen} presents the generative results on the VisDial v1.0 validation split. Our proposed MDST model outperforms all the comparison methods on 4 out of 6 metrics, establishing a new state-of-the-art. Specifically, we achieve an NDCG of 65.03, MRR of 53.49, R@1 of 42.56, and R@5 of 62.47. Notably, when compared to attention- and graph-based methods, our model shows significant improvements across all metrics, especially in NDCG and MRR. We improve NDCG by 2.02 points (65.03 vs. SeqMRN's 63.01) and MRR by approximately 2.17 points (53.49 vs. LTMI-GoG's 51.32). When compared to concatenation-based methods, our MDST model achieves similar or better results. Moreover, it surpasses the previous best-performing method, UTC, by approximately 1.27 points in MRR and 1.17 points in NDCG. It's important to note that UTC relies on ViLBERT pretraining and utilizes VQA datasets.

Table~\ref{tab_test_gen} displays the results on the VisDial v1.0 test split. Our MDST model achieves a NDCG value of 63.18, outperforming other methods across various metrics. Compared to UTC, MDST improves NDCG by 0.53 points. In summary, our MDST model, despite being simpler and not relying on larger pre-trained language models or extra datasets like UTC and ICMU, achieves significant improvements across most metrics, outperforming previous state-of-the-art models. These improvements highlight the effectiveness of the dialogue state tracking mechanism in VisDial.

\noindent {\bf Ablation Studies} Table~\ref{tab_ablation} presents the results of ablation studies, which evaluate the importance of each module in the generative setting. The first row represents the performance of the full model, while the subsequent rows (2-4) indicate the effect of removing each module sequentially.

When removing the QGDS\&PDS module (Row 2), we adopt a similar approach to previous work, where we directly use the question to attend to image features and original history embeddings. The fusion of the question, attended image features, and history features is then fed into the ADer module. The results show a significant drop in NDCG by -3.88 points, MRR decreases to 50.80 by -2.69 points, and R@{1,5,10} exhibits a substantial decrease. These findings align with the previous comparisons and further demonstrate the effectiveness of the dialogue state tracking mechanism. It highlights the ability of the model to capture information flows in the dialogue history at the round level. It is important to note that utilizing PDS without QGDS or vice versa is meaningless since the dialogue states are updated in PDS but used in QGDS. The combination of the QGDS and PDS modules provides strong support for the tracking mechanism.

When removing the switching probability $\alpha^{(t)}$ in the QGDS module, we observe a significant decrease in overall performance. NDCG and MRR decrease by 1.39 and 0.7 points, respectively. This result underscores the importance of the switching probability in our model. Specifically, the switching probability plays a crucial role in associating the two alignment distributions ($\pi^{(t)}{l}$ and $\pi^{(t)}{v}$), facilitating the alignment of vision-language dialogue states. In other words, aligning vision-language states brings about a substantial improvement, which aligns with findings from previous studies \cite{MVAN,AlignVD}.

Furthermore, when removing the two pseudo-object features, NULL and ALL (Row 4), we observe a slight decline in performance. This finding validates that both pseudo-objects carry useful information about the image. The inclusion of these pseudo-objects is valuable because the upcoming question may be unrelated to the input image or may involve the entire image.

\begin{figure*}[!htb]
\centering
\includegraphics[width=0.95\textwidth]{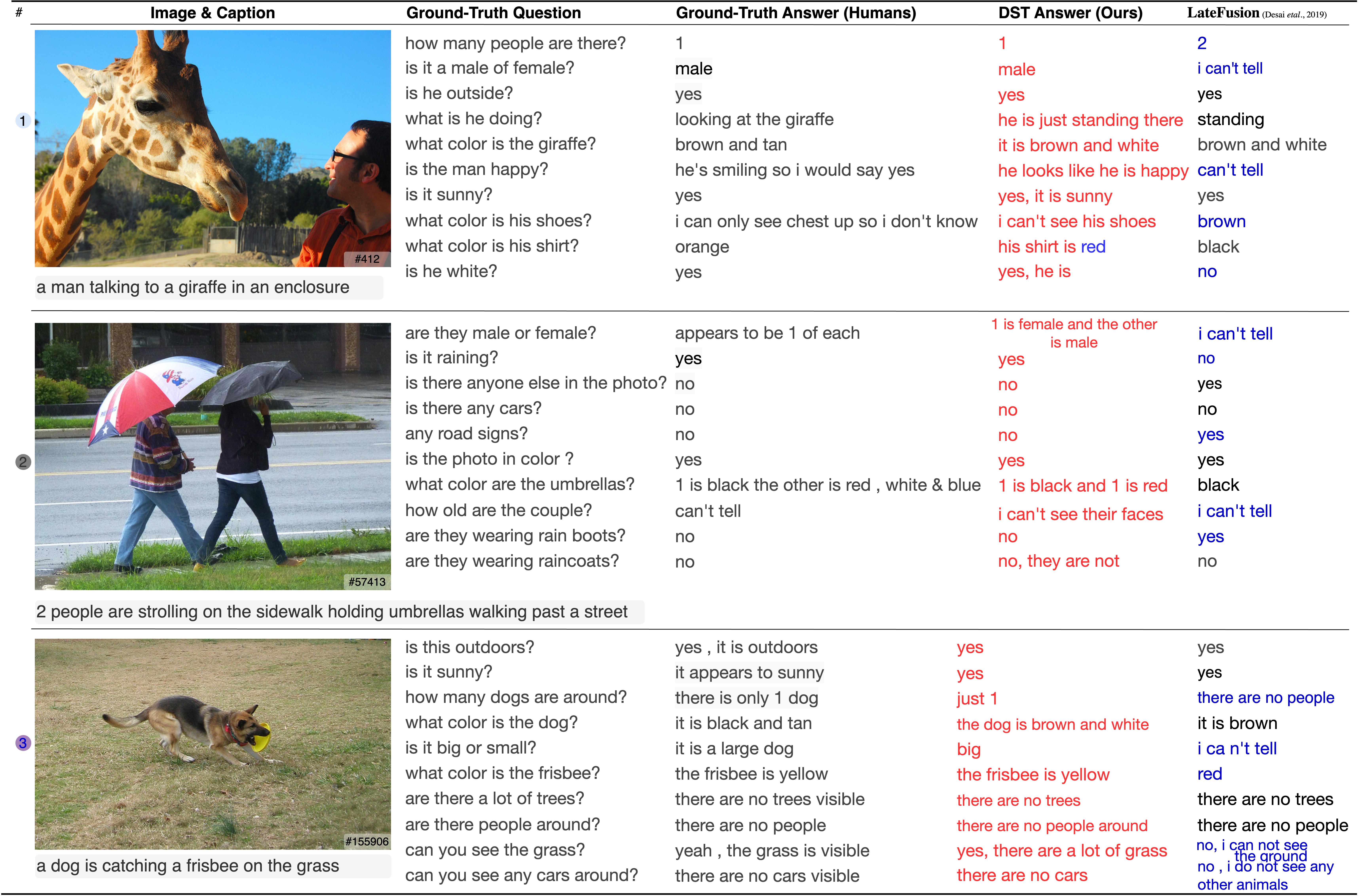}
\caption{Answers generated by our MDST model. The correct answers are highlighted in red, and blue highlights denote the incorrect answers.}
\label{fig_cases}
\end{figure*}
\begin{table}[!htb]\small
\centering
\caption{JACC (\%) and AvgLen on VisDial v1.0 val.} \label{tab_JACC}
\setlength{\tabcolsep}{2.8mm}{
\begin{tabular}{l|c|cc|cc}
\hline
Model& \#QA pair& Correct  &InCorrect& JACC&AvgLen\\
\hline
MDST&\multirow{3}{*}{1000}&798&202&\textbf{79.8}&\textbf{3.57} \\
LateFusion&&534&466&53.4&1.81\\
Human&&-&-&-&3.11\\
\hline
\end{tabular}}
\end{table}

\noindent {\bf Human Studies}  A series of human studies were conducted on VisDial v1.0 val to generate 2064 dialogues. The results are presented in Table~\ref{tab_JACC}, and we provide three examples of generated dialogues in Fig.~\ref{fig_cases}. The findings reveal that our proposed MDST model demonstrates the ability to consistently provide correct answers throughout a series of questions, while generating more human-like responses. Specifically, MDST achieves a Joint Answer Accuracy (JACC) of 79.8\% on VisDial v1.0 val split, indicating that approximately 8 out of 10 rounds yield correct answers. In comparison, LateFusion achieves a significantly lower JACC of 53.4\%.

In the first example, MDST generates 9 correct answers. Notably, in the 6th and 8th rounds, MDST produces reasonable responses: “he looks like he is happy” and “I can't see his shoes”, which capture the semantics similar to the ground truth: “he's smiling so I would say yes” and “I can only see chest up so I don't know”. In the second example, MDST provides correct answers in all 10 rounds, with the response “1 is female and the other is male” being as accurate and natural as the human-generated answer "appears to be 1 of each”. The third example also showcases MDST's ability to correctly predict all 10 questions, with the response “the dog is brown and white” in the 4th round also deemed correct based on the image.

Interestingly, contrary to the human and LateFusion models, MDST tends to produce longer, more consistent answers. The average answer length (AvgLen) of MDST reaches 3.57, surpassing the human-generated answer length of 3.11 and LateFusion's length of 1.81. In terms of consistency, for the question “how many dogs are around?” in the second example, our model responds with an accurate answer “just 1”, which aligns with dialog history (i.e., image caption). In contrast, LateFusion provides an inconsistent response of “there are no people”, and in the last round, LateFusion produces a question-irrelevant answer.

\section{Conclusions}
In this paper, we introduce a novel approach called Multi-Round Dialogue State Tracking Network (MDST) for the task of Visual Dialog (VD). Unlike previous methods that treat dialog history as a simple text input, MDST tracks and updates dialogue states, which are 2-tuple aligned vision-language representations. By modeling the inherent interactions at the round level, MDST aims to capture dynamics of the conversation more effectively. Experimental results on VisDial v1.0 dataset demonstrate that MDST achieves state-of-the-art performance across most evaluation metrics. Additionally, extensive human studies further validate MDST can generate long, consistent, and human-like answers while maintaining the ability to provide correct responses to a series of questions. Overall, our proposed MDST framework represents a significant advancement in visual dialog systems, showcasing the importance of modeling dialogue states in capturing the complex nature of visual conversations. 

\section*{Acknowledgements}
We thank the reviewers for their comments and suggestions. This paper was partially supported by the National Natural Science Foundation of China (NSFC 62076032), Huawei Noah’s Ark Lab, MoECMCC “Artificial Intelligence” Project (No. MCM20190701), Beijing Natural Science Foundation (Grant No. 4204100), and BUPT Excellent Ph.D. Students Foundation (No. CX2020309).

%
%
%
\bibliographystyle{splncs04}
\bibliography{bib_150}
\end{document}